\documentclass[10pt, a4paper]{article}
\usepackage{lrec}
\usepackage{graphicx}
\usepackage{tabularx}
\usepackage{soul}
\usepackage{xcolor}
\usepackage{pdfpages}

\def\url#1{\texttt{#1}}

\usepackage{xstring}
\usepackage{epstopdf}
\usepackage[utf8]{inputenc}

\def\footurl#1{\footnote{\url{#1}}}

\usepackage[noabbrev,capitalize]{cleveref}

\def\parcite#1{\cite{#1}}     
\def\perscite#1{\newcite{#1}} 

\title{
COSTRA 1.0: A Dataset of Complex Sentence Transformations
}

\name{Petra Baran\v{c}\'{i}kov\'{a}, Ond\v{r}ej Bojar}

\address{Charles University \\
         Faculty of Mathematics and Physics\\
         Institute of Formal and Applied Linguistics \\
         \{barancikova,bojar\}@ufal.mff.cuni.cz\\}

\abstract{
We present COSTRA 1.0, a dataset of complex sentence transformations. The dataset is intended for the study of sentence-level embeddings beyond simple word alternations or standard paraphrasing. This first version of the dataset is limited to sentences in Czech but the construction method is universal and we plan to use it also for other languages.
\\
The dataset consists of 4,262 unique sentences with an average length of 10 words, illustrating 15 types of modifications, such as simplification, generalization, or formal and informal language variation. The hope is that with this dataset, we should be able to test semantic properties of sentence embeddings and perhaps even to find some topologically interesting ``skeleton'' in the sentence embedding space. A preliminary analysis using LASER, multi-purpose multi-lingual sentence embeddings suggests that the LASER space does not exhibit the desired properties.
 \\ \newline  }

\begin{document}

\maketitleabstract

\section{Introduction}

Vector representations are essential in the majority of natural language processing tasks. 
The popularity of~word embeddings started with the introduction of word2vec \cite{Mikolov13a} and GloVe \cite{glove} and their properties have been analyzed at length from various aspects. 

Studies of word embeddings range from word similarity \cite{HillRK14,faruqui-dyer-2014-community}, over the ability to capture derivational relations \cite{derinet-word-embeddings}, linear superposition of multiple senses \cite{SuperArora}, the ability to predict semantic hierarchies 
\cite{fu-etal-2014-learning} or POS tags \cite{POS-word-embeddings} up to data efficiency 
\cite{JastrzebskiLC17}.

\begin{table*}[t]
\centering
\begin{tabular}{l|l|r}

\textbf{Change} & \textbf{Example of change} & \textbf{\%} \\
\hline
change of aspect & Hunters have fallen asleep on a clearing. & 4 \\

opposite/shifted meaning & On a clearing, several hunters were dancing. & 15 \\

less formally & Several deer stalkers kipped down on a clearing. & 6 \\

change into possibility &  Several hunters can sleep on a clearing. & 4 \\

ban & Hunters are forbidden to sleep on a clearing. & 4 \\

exaggeration & Crowds of hunters slept on a clearing. & 7 \\

concretization & Several hunters dozed off after lunch on the Upper clearing. & 15 \\

generalization & There were several men in a forest. & 9 \\

change of locality & Several hunters slept in a cinema. & 3 \\

change of gender & Several huntresses slept on a clearing. & 2 \\
\hline
Total &  & 65 \\

\end{tabular}
\caption{Examples of transformations given to annotators for the source sentence 
\textit{Several hunters slept on a clearing.} The third column shows how many of all the transformation suggestions collected in the first round closely mimic the particular example. The number is approximate as annotators typically call one transformation by several names, e.g. \textit{less formally}, 
\textit{formality diminished}, \textit{decrease of formality}, \textit{not formal expressions}, 
\textit{non-formal}, \textit{less formal}, \textit{formality decreased}, ...} The examples were translated to English for presentation purposes only.
\label{tab:first_round_examples}
\end{table*}

Several studies \parcite{mikolov-etal-2013-linguistic,Mikolov13b,levy-goldberg-2014-linguistic,VylomovaRCB15} show that word vector representations are capable of~capturing meaningful syntactic and semantic regularities. These include, for example, male/female relation demonstrated by the pairs ``man:woman'', ``king:queen'' and the country/capital relation (``Russia:Moscow'', ``Japan:Tokyo''). These regularities correspond to simple arithmetic operations in the vector space.

Sentence embeddings are becoming equally ubiquitous in~NLP, with novel representations appearing almost every other week. With an overwhelming number of methods to~compute sentence vector representations, the study of their general properties becomes  difficult. Furthermore, it is not entirely clear in which way the embeddings should be evaluated.

In an attempt to bring together more traditional representations of sentence meanings and the emerging vector representations,
\perscite{bojar:etal:jnle:representations:2019} introduce several aspects or desirable properties of sentence embeddings. One of~them, ``relatability'', highlights the correspondence of meaningful differences between sentences on the one hand and geometrical relations between their respective embeddings in the highly dimensional continuous vector space on the other hand.
If we found such correspondence, we could apply geometrical operations in the space to induce meaningful changes in sentences.

In this work, we present COSTRA, a new dataset of COmplex Sentence TRAnsformations. In its first version, the dataset is limited to sample sentences in Czech. The goal is to support studies of semantic and syntactic relations between sentences in the continuous space. Our dataset is the prerequisite for one of the possible ways of exploring sentence meaning relatability:\footnote{The term ``relatability'' is used to indicate that we search for specific types of relations among sentences. The common term ``relatedness'', in our opinion, suggests some vagueness on the relation type. We do not build a dataset of sentences related in just some way, we seek for a set of clear-cut, ``orthogonal'' relations.} We envision that the continuous space of sentences induced by an ideal embedding method would exhibit topological similarity to the graph of sentence variations. For instance, one could argue that a subset of sentences could be organized along a linear scale reflecting the formalness of the language used. Another set of sentences could form a partially ordered set of gradually less and less concrete statements. And yet another set, intersecting both of the previous ones in multiple sentences could be partially or linearly ordered according to the strength of the speaker's confidence in the claim.

Our long term goal is to search for a sentence embedding method that exhibits this behaviour, i.e., that the topological map of~the embedding space corresponds to meaningful operations or changes in the set of sentences of a language (or more languages at once). We prefer this behaviour to \emph{emerge}, as it happened for word vector operations, but regardless if the behaviour is emergent or trained, we need a dataset of sentences illustrating these patterns. A large dataset could serve for training; a small one would provide a test set. In either case, these sentences could provide a ``skeleton'' to the continuous space of sentence embeddings.\footnote{The Czech word for
``skeleton'' is ``kostra''.}

The paper is structured as follows: \cref{related} summarizes existing methods of sentence embeddings evaluation and related work. 
\cref{annotation} describes our methodology for constructing our dataset. \cref{data} details the obtained dataset and some first observations. We conclude and provide the link to the dataset in \cref{conclusion}

\section{Background}
\label{related}


As hinted above, there are many methods of converting a sentence into a vector in a highly dimensional space. To~name a~few: BiLSTM with the max-pooling trained for natural language inference \cite{infersent}, masked language modelling and next sentence prediction using  bidirectional Transformer \cite{bert}, max-pooling last states of neural machine translation among many languages  \cite{laser}  or the encoder final state in attentionless neural machine translation
\parcite{cifka:bojar:meanings:2018}.

The most common way of evaluating methods of sentence embeddings is extrinsic, using so-called `transfer tasks', i.e., comparing embeddings via the performance in downstream tasks such as paraphrasing, entailment, sentence sentiment analysis, natural language inference and other assignments. However, even simple bag-of-words (BOW) approaches often achieve competitive results on such tasks \cite{wieting2016iclr}.


\newcite{Adi16} introduce intrinsic evaluation by measuring the ability of models to encode basic linguistic properties of a sentence such as its length, word order, and word occurrences.
These so-called `probing tasks' are further extended by a depth of the syntactic tree, top constituent or verb tense by \newcite{DBLP:journals/corr/abs-1805-01070}.

Both transfer and probing tasks are integrated into SentEval \cite{senteval} framework for sentence vector representations. \perscite{Perone2018} applied SentEval to eleven different encoding methods revealing that there is no consistently well-performing method across all tasks. SentEval was further criticized for pitfalls such as comparing different embedding sizes or correlation between tasks \cite{pitfalls2019,DBLP:journals/corr/abs-1901-10444}.

\perscite{shi-etal-2016-string} show that NMT encoder is able to capture syntactic information about the source sentence. \perscite{DBLP:journals/corr/BelinkovDDSG17} examine the ability of
NMT to learn morphology through POS and morphological tagging.

\begin{table*}[t]
\begin{center}
\begin{tabular}{l|l}
   \textbf{Change} & \textbf{Instructions} \\
   \hline
   paraphrase 1 & Reformulation the sentence using different words \\

   paraphrase 2 & Reformulation the sentence using other different words \\

   different meaning & Shuffle words in the sentence in order to get different meaning\\
  
   opposite meaning & Reformulate the sentence to get a sentence with opposite meaning\\
 
   nonsense & Shuffle words in sentence to make grammatical sentence with no sense. \\
   & E.g. \textit{A hen pecked grain.} $\to$ \textit{Grain pecked a hen.} \\

   minimal change & Significantly change the meaning of the sentence using only a minimal alternation.\\

   generalization & Make the sentence more general. \\
 
   gossip & Rewrite the sentence in a gossip style -- strongly exaggerated meaning on the sentence. \\
 
   formal sentence & Rewrite the sentence in a more formal style. \\
  
   non-standard sentence & Rewrite the sentence in non-standard, colloquial style. \\
   
   simple sentence & Rewrite the sentence in a simplistic style, so even a person with a limited vocabulary \\
   & could understand it. \\

   possibility & Change the modality of the sentence into a possibility. \\

   ban & Change the modality of the sentence into a ban. \\
 
   future & Move the sentence into the future. \\

   past & Move the sentence into the past. \\

\end{tabular}
\caption{Sentences transformations requested in the second round of annotation with the instructions to the annotators. The annotators were given no examples
(with the exception of \textit{nonsense}) not to be influenced as much as in the first round.}
\label{annotation_instructions} 
 \end{center}
\end{table*}

Still, very little is known about the semantic properties of~sentence embeddings. 
Interestingly, \perscite{cifka:bojar:meanings:2018} observe that the better self-attention embeddings serve in NMT, the worse they perform in most of SentEval tasks.

\newcite{zhu-etal-2018-exploring}  generate automatically sentence variations such as: 
\begin{itemize}
    \item[(1)] Original sentence: \textit{A rooster pecked grain.}
    \item[(2)] Synonym Substitution: \textit{A cock pecked grain.}
    \item[(3)] Not-Negation: \textit{A rooster didn't peck grain.}
    \item[(4)] Quantifier-Negation: \textit{There was no rooster pecking grain.} 
\end{itemize}

and compare their triplets by examining distances between their embeddings, i.e. distance between (1) and (2) should be smaller than distances between (1) and (3), (2) and (3), similarly, (3) and (4) should be closer together than (1)--(3) or (1)--(4).

In our previous study \cite{BaBo2019}, we examined the effect of small sentence alternations in sentence vector spaces. 
We used sentence pairs automatically extracted from datasets for natural language inference SNLI \cite{snli} and MultiNLI \cite{MultiNLI}. We observed that the vector difference, familiar from word embeddings, serves reasonably well also in~sentence embedding spaces. 
The examined relations were, however, very simple: a change of gender, number, the addition of an adjective, etc. 
The structure of the sentence and its wording remained almost identical.

We would like to move to more interesting non-trivial sentence comparison, beyond those in \perscite{zhu-etal-2018-exploring} or \perscite{BaBo2019}, such as change of style of~a~sentence, the introduction of a small modification that drastically changes the meaning of a sentence or reshuffling of words in a sentence so that its meaning is altered.

Unfortunately, such a dataset cannot be generated automatically and it is not available to our best knowledge. We attempt to start filling this gap with COSTRA 1.0.

\section{Annotation}
\label{annotation}

We acquired the data in two rounds of annotation. 
In the first one, we were looking for original and uncommon sentence change suggestions. 
In the second one, we collected sentence alternations using ideas from the first round. 
The first and second rounds of annotation could be broadly called as \textit{collecting ideas} and \textit{collecting data}, respectively.

\subsection{First Round: Collecting Ideas}
We manually selected 15 newspaper headlines. 
Eleven annotators were asked to modify each headline up to 20 times and describe the modification with a short\footnote{This requirement was not always respected. The annotators sometimes created very complex descriptions such as \textit{specification of information about the society affected by the presence of an alien}.} name. 
They were given an example sentence and several of its possible alternations,
see \cref{tab:first_round_examples} on the preceding page.

Unfortunately, these examples turned out to be highly influential on the annotators' decisions and they correspond to almost two-thirds of all of the modifications gathered in~the first round. 
Other very common transformations include change of word order or transformation into an interrogative/imperative sentence.

Other suggested interesting alterations include change into a fairy-tale style,  excessive use of diminutives/vulgarisms or dadaism---a swap of roles in the sentence so that the resulting sentence is grammatically correct but nonsensical in our world. 
Of these suggestions, we selected only the dadaistic swap of roles for the current exploration (see \textit{nonsense} in Table \ref{annotation_instructions}).

In total, we collected 984 sentences with 269 described unique changes. 
We use them as inspiration for the second round of annotation.

\subsection{Second Round: Collecting Data}

\paragraph{Sentence Transformations}

We selected 15 modifications types to collect COSTRA 1.0. They are presented in~\cref{annotation_instructions}.  

We asked for two distinct paraphrases of each sentence because we believe that a proper sentence embedding should put paraphrases close together in vector space. 

Several modification types were explicitly selected to constitute a thorough test of embeddings. 
In \textit{different meaning}, the annotators should create a sentence with some other meaning using the same words as the original sentence.
Other transformations that should be challenging for embeddings include \textit{minimal change}, in which the sentence meaning should be significantly modified by only minimal alternation, or \textit{nonsense}, in which words of the source sentence should be rearranged into a grammatically correct sentence without any sense.

\begin{table*}[t]
\centering
\begin{tabular}{l|r|r|r|r|r|r}
\bf Annotator     
& \textbf{ \# Annotations} 
& \textbf{ \# Sentences} 
& \textbf{ \# Impossible} 
& \textbf{ \# Typos} 
& \textbf{Avg. Sent. Length} 
& \textbf{Avg. Time}  \\
\hline
armadillo    & 69     & 1035  & 0  & 9  & 10.3   & 12:32        \\
wolverine    & 42     & 598   & 32 & 13 & 9.6    & 14:32        \\
honeybadger  & 39     & 584   & 1  & 28 & 10.4   & 30:38        \\
gorilla      & 31     & 448   & 17 & 16 & 9.8    & 16:55        \\
porcupine    & 31     & 465   & 0  & 6  & 11.3   & 8:55         \\
lumpfish     & 23     & 329   & 16 & 4  & 8.4    & 13:28        \\
crane        & 22     & 319   & 11 & 15 & 9.2    & 15:30        \\
meerkat      & 17     & 241   & 14 & 17 & 9.1    & 27:36        \\
axolotl      & 8      & 116   & 4  & 11 & 10.1   & 24:02        \\
bullshark    & 6      & 90    & 0  & 2  & 9.8    & 20:59        \\
flamingo     & 3      & 45    & 0  & 8  & 11.3   & 11:37        \\
capybara     & 2      & 30    & 0  & 0  & 7.6    & 25:06         \\
\hline
Total        & 293    & 4,300 & 95 & 129  & 9.9    & 19:50       \\
\end{tabular}
\caption{Statistics for individual annotators (anonymized as armadillo, \dots{}, capybara).}
\label{tab:statistics}
\end{table*}

\paragraph{Seed Data}

The source sentences for annotations were selected from the Czech data of Global Voices \cite{tiedemann-2012-parallel} and OpenSubtitles\footurl{http://www.opensubtitles.org/} \cite{opensubtitles2016}. 
We used two sources in order to have different styles of seed sentences, both journalistic and common spoken language. 
We considered only sentences with more than 5 and less than 15 words and we manually selected 150 of them for further annotation. 
This step was necessary to remove sentences that are:

\begin{itemize}
    \item too unreal, out of this world, such as: \\
          \textit{Jedno fotonov\'{y} torp\'{e}do a je z tebe vesm\'{i}rn\'{a} topinka.} \\
          ``One photon torpedo and you're a space toast.''
    \item photo captions (i.e. incomplete sentences), e.g.: \\
          \textit{Zvl\'{a}\v{s}tn\'{i} ekv\'{a}dorsk\'{y} p\v{r}\'{i}pad Correa vs. Crudo} \\
          ``Specific Ecuadorian case Correa vs. Crudo''
    \item too vague, overly dependent on the context: \\
          \textit{B\v{e}\v{z} tam a mluv na ni.} \\
          ``Go there and speak to her.''
\end{itemize}

Many of the intended sentence transformations would be impossible to apply to such sentences and annotators' time would be wasted.
Even after such filtering, it was still quite possible that the desired sentence modification could not be achieved for a sentence.
For such a case, we gave the annotators the option to enter the keyword \textit{IMPOSSIBLE} instead of the particular (impossible) modification.

This option allowed to state that no such transformation is possible explicitly. At the same time, most of the transformations are likely to lead to a large number of possible outcomes.
As documented in \perscite{scratching2013}, a Czech sentence might have hundreds of thousands of paraphrases. To support some minimal exploration of this possible diversity, most of the sentences were assigned to several annotators. 

\paragraph{Spell-Checking}
The annotation is a challenging task and the annotators naturally make mistakes. 
Unfortunately, a~single typo can significantly influence the resulting embedding \cite{malykh-etal-2018-robust}. 
After collecting all the sentence variations, we applied the statistical spellchecker and grammar checker 
Korektor \cite{richter12} in order to minimize influence of typos to performance of embedding methods. 
We manually inspected 519 errors identified by Korektor and fixed 129 true errors.

\section{Dataset Description}
\label{data}

In the second round, we collected 293 annotations from 12 annotators. 
After Korektor, there are 4262 unique sentences (including 150 seed sentences) that form the COSTRA 1.0 dataset. 
Statistics of individual annotators are available in \cref{tab:statistics}.

The time needed to carry out one piece of annotation (i.e., to provide one seed sentence with all 15 transformations) was on average almost 20 minutes but some annotators easily needed even half an hour. 
Out of the 4262 distinct sentences, only 188 were recorded more than once. 
In other words, the chance of two annotators producing the same output string is quite low. 
The most repeated transformations are by far \textit{past}, \textit{future} and \textit{ban}. 
The least repeated is \textit{paraphrase} with only one sentence repeated.

\begin{table}[t]
\begin{center}
\begin{tabular}{l|c|r|r}

     \textbf{Persons} & \textbf{ \# Annotations} & \textbf{Unique Sents. } & \textbf{U.S. \%} \\
      \hline
      1 & 30 & 438 & 99,8\%\\
      2 & 30 & 851 & 97,3\%\\ 
      3 & 61 & 2545 & 94,3\%\\
      4 & 5 &  278  & 95,8\% \\
      \hline
      Total & 126 & 4112 & 96,8\% \\

\end{tabular}
\caption{The number of people annotating the same sentence. Most of the
sentences have at least three different annotators. Unfortunately, 24 sentences
were left without any annotation.}
\label{multiple-annots}
 \end{center}
\end{table}

\cref{multiple-annots} documents this in another way. 
The 293 annotations are split into groups depending on how many annotators saw the same input sentence: 30 annotations were annotated by one person only, 30 annotations by two different persons, etc. 
The last column shows the number of unique outputs obtained in that group. 
Across all cases, 96.8\% of produced strings were unique.\footnote{The number of unique outputs from single-person annotations is not 100\% because one of the annotators wrongly produced the same sentence for both \textit{possibility} and \textit{future} transformation.}

\begin{figure*}[!h]
 
 \includegraphics[width=2\columnwidth]{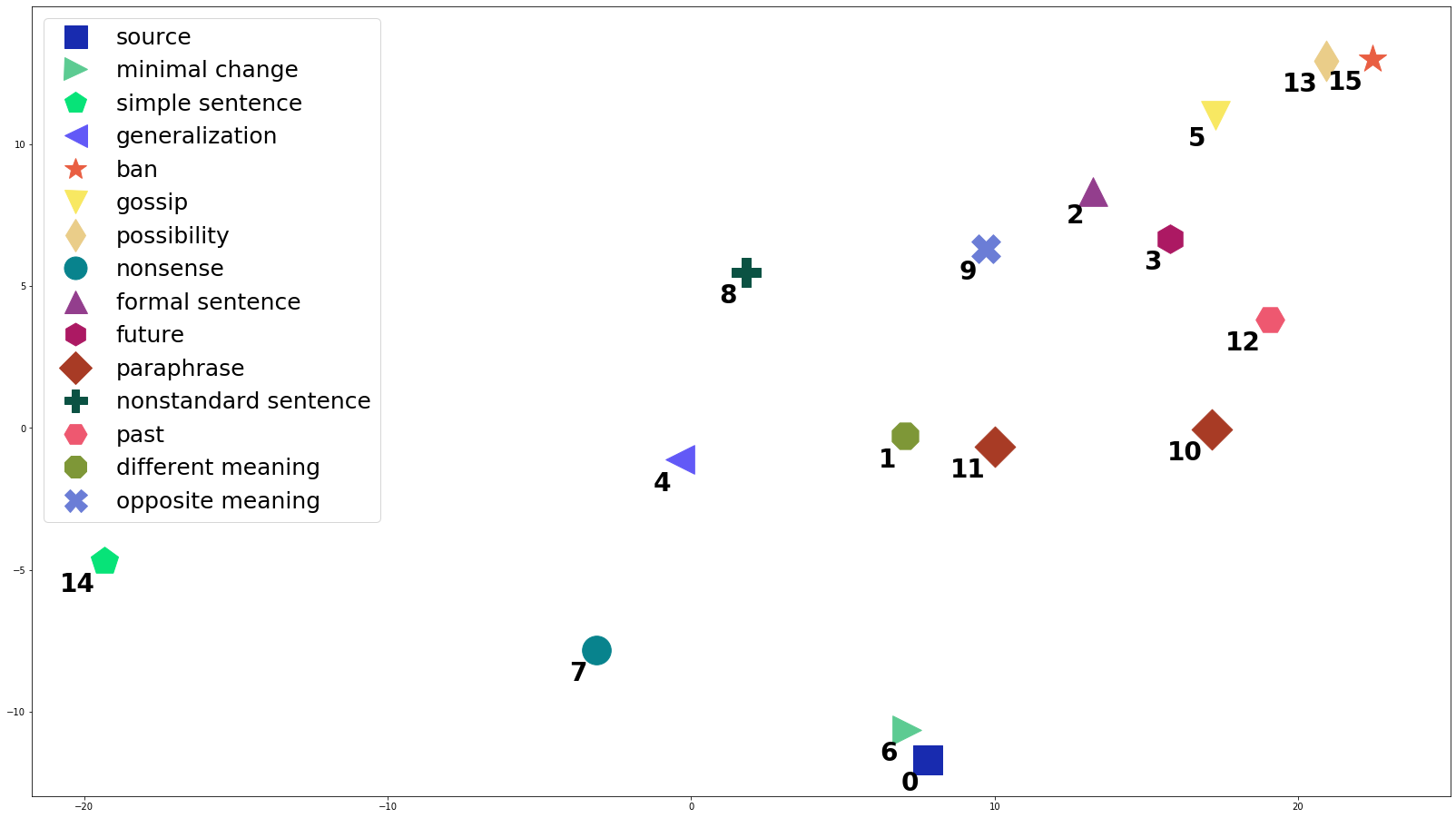}
 \resizebox{\textwidth}{!}{%
    \begin{tabular}{c|l|l}
\textbf{id} & \textbf{change} & \textbf{transformation and its translation} \\
\hline
0 & source & Je to prostě blbost, zrušte ten projekt, prosím, a nemrhejte státními penězi. \\
& & \textit{It's just crap, please cancel the project, and don't waste state money.} \\
1 & different meaning & Zrušení projektu je blbost a nesmyslné mrhání státními penězi. \\
& & \textit{Cancellation of the project is stupid and pointless waste of state money.} \\
2 & formal sentence & Ten projekt je od základu špatný, zrušte jej prosím a neplýtvejte státními prostředky. \\
& & \textit{The project is fundamentally bad, please cancel it and don't waste state resources.} \\
3 & future & Bude to určitě blbost, projekt bude nejspíš zrušen a nebude se mrhat státními penězi. \\
& & \textit{It will certainly be crap, the project will likely be canceled and state money will not be wasted.} \\
4 & generalization & Projekt je blbost, zastavte jej a nemrhejte penězi. \\
& & \textit{The project is crap, stop it and don't waste money.} \\
5 & gossip & Ten projekt je šílenost, měl by se okamžitě zastavit a neplýtvat na něm miliony z našich daní. \\
& & \textit{The project is crazy, it should immediately stop and not to waste millions on it from our taxes.} \\
6 & minimal change & Je to blbost, přerušte ten projekt, prosím, a mrhejte státními penězi. \\
& & \textit{It's crap, interrupt the project, please, and waste state money.} \\
7 & nonsense & Zrušte státní peníze prosím a nemrhejte prostými blbostmi. \\
& & \textit{Please cancel the state money and do not waste plain crap.} \\
8 & nonstandard sent. & Je to kravina, práce by se měla zastavit a nemrhat státníma penězma. \\
& & \textit{It's bullshit, work should stop and not waste government money.} \\
9 & opposite meaning & Jedná se o promyšlený plán, je třeba ho realizovat a uvolnit peníze ze státní kasy. \\
& & \textit{It is a well-thought-out plan, it needs to be implemented and the money released from the Treasury.} \\
10 & paraphrase & Je to hloupost, zastavte ten projekt a ušetřete státní peníze. \\
& & \textit{It's stupid, stop the project and squander state money.} \\
11 & paraphrase & Je to blbost, zastavte práce a neplýtvejte veřejnými prostředky. \\
& & \textit{It's stupid, stop the work and don't waste public funds.} \\
12 & past & Byla to prostě blbost, projekt byl zrušen a nemrhalo se státními penězi. \\
& & \textit{It was just crap, the project was canceled and state money weren't wasted.} \\
13 & possibility & Tento projekt může být blbost, mohl by se zrušit a nemělo by se zde mrhat státními penězi. \\
& & \textit{This project can be stupid, could be canceled and state money shouldn't be wasted.} \\
14 & simple sentence & Projekt je špatný, zastavte jej, neplýtvejte státními penězi. \\
& & \textit{The project is bad, stop it, don't waste state money.} \\
15 & ban &  Ten projekt nesmí být blbost, jeho realizace se nesmí zrušit a nesmí být na něj uvolněny státní peníze.\\
& & \textit{The project mustn't be crap, its realization mustn't be canceled and state money mustn't be released for it.} \\
    \end{tabular}}
    \caption{2D visualization using PCA of a single annotation. 
             Sentences corresponding to the numbers in the plot are listed under the visualization. 
             Best viewed in colors.}
     \label{fig:pca}
\end{figure*}

In line with instructions, the annotators were using the \textit{IMPOSSIBLE} option scarcely (95 times, i.e., only 2\%). 
It was also a case of 7 annotators only; the remaining 5 annotators were capable of producing all requested transformations. 
The top three transformations considered unfeasible were \textit{different meaning} (using the same set of words), \textit{past} (esp. for sentences already in the past tense)\footnote{The annotators clearly did not consider the option to express a more distant past lexically.}  and \textit{simple sentence}.

\paragraph{First Observations}
We embedded COSTRA sentences with LASER \cite{laser}, the only currently available off-the-shelf sentence embedding model for the Czech language.
Having browsed a number of 2D visualizations (PCA and t-SNE) of the space, we have to conclude that visually, LASER space does not seem to exhibit any of the desired topological properties discussed above, see \cref{fig:pca} for one example.

\cref{similarities} summarizes vector and string similarities between seed sentences and their transformations. It reflects the lack of semantic relations in the LASER space -- the embedding of \textit{minimal change} transformation lies very close to the original sentence (average similarity of 0.945) even though the transformation substantially changed the meaning of the sentence. Tense changes and some form of negation or banning also keep the vectors very similar.

The lowest average similarity was observed for \textit{generalization} (0.739) and \textit{gossip} (0.809), which is not any bad sign. 
However the fact that \textit{paraphrases} have much smaller similarity (0.827) than \textit{opposite meaning} (0.902) documents that the vector space lacks in terms of ``relatability''.

The string similarity between two sentences $s_1$ and $s_2$ was computed as $\frac{|s_1|+|s_2|-d_L(s_1,s_2)}{|s_1|+|s_2|}$, where $d_L$ represents Levenshtein distance.
Pearson correlation of the average cosine similarity between seed sentence embeddings and their transformation and  average string similarity is 0.934, i.e., very strong correlation. 
This result suggests that LASER embeddings are superficial and lack a deeper grasp into the meaning of sentences.

\begin{table}[t]
\begin{center}
\resizebox{\columnwidth}{!}{%
\begin{tabular}{l|r|r}
\bf Transformation  & \bf Vector Similarity & \bf String Similarity \\
\hline
minimal change       & 0.945 & 0.887 \\
past                 & 0.915 & 0.864 \\
future               & 0.909 & 0.859 \\
opposite meaning     & 0.902 & 0.821 \\
possibility          & 0.899 & 0.843 \\
ban                  & 0.895 & 0.819 \\
nonsense             & 0.881 & 0.675 \\
different meaning    & 0.869 & 0.699 \\
nonstandard sentence & 0.851 & 0.660 \\
formal sentence      & 0.850 & 0.661 \\
paraphrase           & 0.827 & 0.556 \\
simple sentence      & 0.810 & 0.606 \\
gossip               & 0.809 & 0.562 \\
generalization       & 0.739 & 0.512 \\
\end{tabular}}
\end{center}
\caption{Vector and string similarity between seed sentences and their transformations per category measured as average cosine similarity and average Levenshtein similarity, respectively.}
\label{similarities}
\end{table}

\section{Conclusion and Future Work}
\label{conclusion}

We presented COSTRA 1.0, a small corpus of complex transformations of Czech sentences.

We plan to use this corpus to analyze a broad spectrum sentence embeddings methods to see to what extent the continuous space they induce reflects semantic relations between sentences in our corpus. 
The very first analysis using LASER embeddings indicates a lack of ``meaning relatability'', i.e., the ability to move along a trajectory in the space in order to reach desired sentence transformations.
Actually, not even paraphrases appear in close neighbourhoods of embedded sentences. 
More ``semantic'' sentence embeddings methods are thus to be sought for.

The corpus is freely available at the following link:
\begin{center}
\url{http://hdl.handle.net/11234/1-3123}
\end{center}

Aside from extending the corpus in Czech and adding other language variants, we are also planning to wrap COSTRA 1.0 into an API such as SentEval making the evaluation of sentence embeddings in terms of ``relatability'' effortless.

\section{Acknowledgment}
The research reported in this paper has been supported by the grant No.
19-26934X (NEUREM3) of the Czech Science Foundation and by the grant No. 825303 (Bergamot) of European Union’s Horizon 2020 research and innovation programme.

This work has been using language resources stored and distributed by the project No.\ LM2015071, \textit{LINDAT-CLARIN}, of the Ministry of Education, Youth and Sports of the Czech Republic.

\section{Bibliographical References}
\label{main:ref}

\bibliographystyle{lrec}
\bibliography{biblio}


\end{document}